\documentclass[sigconf,natbib=false,nonacm]{acmart}

\AtBeginDocument{%
  }

\setcopyright{acmcopyright}
\copyrightyear{2023}
\acmYear{2023}

\acmSubmissionID{27}
\RequirePackage[
  datamodel=acmdatamodel,
  style=acmnumeric,
  ]{biblatex}
  
\usepackage{algorithm}
\usepackage{algpseudocode}

\begin{document}

\title{Hybrid Models for Facial Emotion Recognition in Children }

\author{Rafael Zimmer}
\affiliation{%
  \institution{University of São Paulo}
  \country{Brazil}
}\email{rafael.zimmer@usp.br}

\author{Marcos Sobral}
\affiliation{%
  \institution{Federal University of Tocantins}
  \country{Brazil}
}\email{marcos.lima2@estudante.ifto.edu.br}

\author{Helio Azevedo}
\affiliation{%
  \institution{Renato Archer Information Technology Center}
  \country{Brazil}
}\email{hazevedo.cti@gmail.com}

\begin{abstract}
    This paper focuses on the use of emotion recognition techniques to assist psychologists in performing children´s therapy through remotely robot operated sessions. In the field of psychology, the use of agent-mediated therapy is growing increasingly given recent advances in robotics and computer science. Specifically, the use of  Embodied Conversational Agents (ECA) as an intermediary tool can help professionals connect with children who face social challenges such as Attention Deficit Hyperactivity Disorder (ADHD), Autism Spectrum Disorder (ASD) or even who are physically unavailable due to being in regions of armed conflict, natural disasters, or other circumstances. In this context, emotion recognition represents an important feedback for the psychotherapist. In this article, we initially present the result of a bibliographical research associated with emotion recognition in  children. This research revealed an initial overview on algorithms and datasets widely used by the community. Then, based on the analysis carried out on the results of the bibliographical research, we used the technique of dense optical flow features to improve the ability of identifying emotions in children in uncontrolled environments. From the output of a hybrid model of Convolutional Neural Network, two intermediary features are fused before being processed by a final classifier. The proposed architecture was called \textit{HybridCNNFusion}. Finally, we present the initial results achieved in the recognition of children's emotions using a dataset of Brazilian children.
\end{abstract}

\maketitle

\section{Introduction}
Human cognitive development goes through several stages from birth to maturity. Childhood represents the phase where one acquires the basis of learning to relate with others and with the world \cite{piaget1952origins}. Unfortunately, the mental development process of a child can be hampered by mental disorders such as anxiety, stress, obsessive-compulsive behavior or emotional, sexual or physical abuse \cite{weisz2010evidence}.
The solution or reduction of consequences for these afflictions is achieved with therapeutic processes carried out by professionals in the field of psychology. Due to limited child maturity, the process involves not only assessment sessions with the child, but also interviews with parents and educators, observation of the child in the residential and school environments and data collection through drawings, compositions, games and other activities \cite{Cuadrado2019}, \cite{WOS:000629356500006}.

In this process, leisure resources such as: games, theater activities, puppets, toys and others gain special prominence and are used as support in therapy \cite{Moura2000}. As a way to contribute to this approach, Embodied Conversational Agents (ECA) are used as a tool in psycho-therapeutic applications. Provoost et al. \cite{Provoost2017} performed a scoping review on the use of ECAs in psychology. After selection, the search revealed 49 references associated with the following mental disorders: autism, depression, anxiety disorder, post-traumatic stress disorder, psychotic disorder and substance use. According to the authors, "ECA applications are very interesting and show promising results, but their complex nature makes it difficult to prove that they are effective and safe for use in clinical practice". Actually, the strategy suggested by Provoost et al.  involves increasing the evidence base through interventions using low-technology agents that are rapidly developed, tested, and applied in responsible clinical practice.

The recognition of emotions during psycho-therapeutic sessions can act as an aid to the psychology professional involved in the process, with a still big room for improvement considering the depth of the task at hand \cite{WOS:000556121100052}. 

The objective of this work is to discuss and comment on the use of images generated by cameras in uncontrolled children's psychotherapy sessions to classify their emotional state at any given moment in one of the following basic emotion categories: anger, disgust, fear, happiness, sadness, surprise, contempt \cite{ekman1984}. Given the diversity of Machine Learning algorithms for emotion recognition tasks overall, correctly addressing our objective is much more complex than simply choosing the most powerful or recent algorithm \cite{WOS:000483067100012}. For applications in psychology, compared to other human-centered tasks, the solution has to be almost fail-proof and be able to function in real uncontrolled scenarios, which is in itself extremely challenging and therefore raises multiple ethical and morally debatable questions about the viability of such models \cite{WOS:000697827200029}. In this context, it is important to study and consider the environments for which a specific algorithm will be used for even before beginning to develop or train it \cite{WOS:000447336700033}. 

In Section \ref{sec:Bibliography}, we briefly discuss the performed bibliographical research on the state of the art for emotion recognition in children. The training datasets, as well as the implemented model architecture and produced code are presented in Sections \ref{subsec:Datasets} and \ref{subsec:Architecture} respectively. Results obtained using the suggested model and conclusions are discussed in Sections \ref{sec:Results} and \ref{sec:Conclusion}.

\section{ Bibliographical Research }
\label{sec:Bibliography}
A bibliographical research was performed to determine the State of The Art (SOTA) for emotion recognition tasks (FER) in children using computer algorithms. The search was made using the ”Web of Science” repository \cite{WEB2023}, covering the last 5 years, with the following search key: 
\begin{equation}
\begin{split}
    \text{\textit{child* AND emotion AND (recognition OR detection) AND}} \\
    \text{\textit{(algorithm OR "machine learning" OR "computer vision")}}
\end{split}    
\end{equation}

An initial number of 152 references were selected, with a total of 42 accepted for in-depth reading (39 from the original search, and 3 additional references). A further reading analysis was done, by tagging each paper according to a select number of categories, including, but not limited to: datasets used; age of the patients; psychological procedure adopted; data format (such as video, photos or scans); algorithm category (deep learning techniques, pure machine learning, etc). The detailed result of this categorization can be seen in the spreadsheet available in the Google Drive \cite{SPREED2023}. 

\subsection{ Types of algorithms and datasets}

\begin{figure}[!h]
    \center
    \includegraphics[scale=0.40]{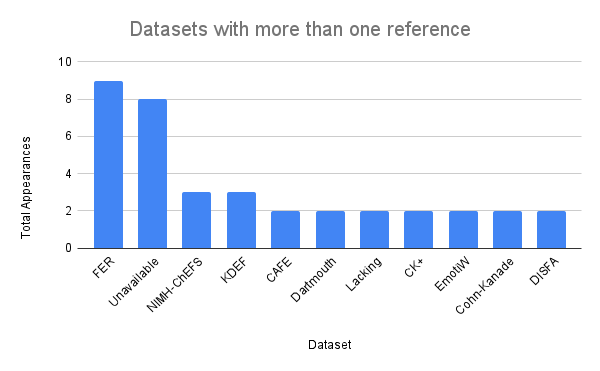}
    \caption{Datasets used for training.}
    \label{fig:datasets}
\end{figure}

In Fig.~\ref{fig:datasets} we present the main datasets identified during the bibliographical research. The FER-2013 dataset \cite{Sambare2022} is one of the most used by researchers with 9 references. We can mention the works by Sreedharan et al. \cite{Sreedharan2018} which makes use of this dataset for training FER model using a novel optimisation technique (Grey Wolf optimisation), for instance.

\begin{figure}[!ht]
    \center
    \includegraphics[scale=0.40]{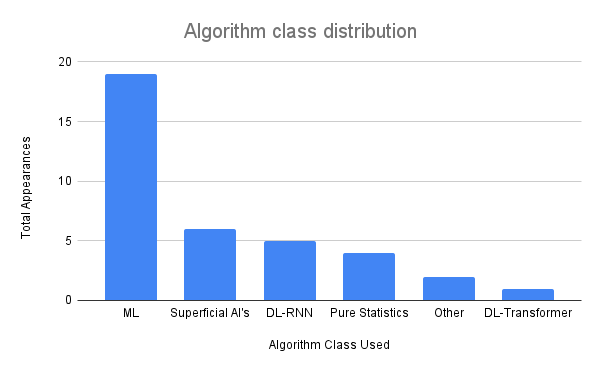}
    \caption{Algorithms for emotion recognition.}
    \label{fig:algorithm_classes}
\end{figure}

Overall, we found out that Facial Emotion Recognition (FER) algorithms have had significant improvement in recent years \cite{WOS:000772182600055}, driven by the success of deep learning-based approaches. In Fig.~\ref{fig:algorithm_classes} we present the most frequently used algorithms for emotion recognition. The convolutional neural network architecture (DL-CNN) was the most used, with 22 references. As DL-CNN examples, we can cite the works of Haque and Valles \cite{Haque2018} and Cuadrado et al.\cite{Cuadrado2019}. Both of these propose an architecture for a Deep Convolutional Neural Network for a specific FER task, namely for robot tutors in primary schools and identifying emotions in children with autism spectrum disorder.

With the demand for high-performance algorithms, numerous novel models, such as the DeepFace system \cite{Taigman2014} or the Transformer architecture for sequential features \cite{vaswani2017attention} have also made great steps in improving the overall accuracy and time efficiency for emotion classification models. 

Among the most popular paradigms currently used for FER, Convolutional Neural Networks (CNNs) have demonstrated high performance in detecting and recognizing emotion features from facial expressions in images \cite{WOS:000697827200029} by applying moving filters over an image, also called convolution kernels. These models use hierarchical feature extraction techniques to construct region-based information from facial images, which are then used for classification. One of the first wide-spread CNN-based models that have been used for FER is the VGG-16 network, which uses 16 convolution layers and 3 fully connected layers to classify emotions \cite{vgg}. In addition to CNNs, other models such as Recurrent Neural Networks (RNNs) or a combination of both have also been proposed for FER. 

Overall, FER is an active area of research, and there is ongoing work to improve the accuracy and robustness of existing solutions.

\subsection{Classic emotion capture strategies}

\begin{figure}[!ht]
    \center
    \includegraphics[scale=0.40]{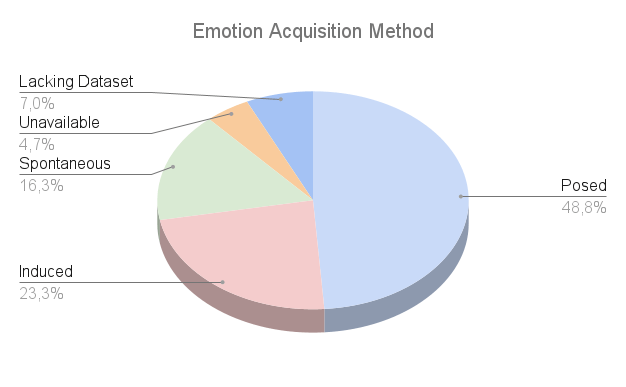}
    \caption{Emotion capture strategies.}
    \label{fig:emotion_acquisition}
\end{figure}

 In Fig.~\ref{fig:emotion_acquisition} we present the origin of the still emotion pictures present in the datasets. We can observe that 48.8\% of the studies used \textit{"Posed"} emotions, such that the emotions expressed are artificial, and their enactment requested by an evaluator. As an example of works that use \textit{"Posed"} emotions we mention Sreedharan et al. \cite{Sreedharan2018} which uses the CK+ dataset of posed emotions, and Kalantarian et. al \cite{Kalantarian2020}, in which children with Autism Spectrum Disorder (ASD) are requested to imitate the emotions shown by prompts in a mobile game. 

The \textit{"Induced"} group of emotions contributes to 23.3\% of the found papers, for which we can mention the work of Goulart et al. \cite{Goulart2019}, where children emotions are induced by interaction with a robot tutor and recorded. Differently from posed emotions, which are obtained by explicitly requesting participants to imitate the facial expression, induced emotions are implicitly obtained by showing the participants emotion inducing scenes, such as videos, photographs and texts.

The \textit{"Spontaneous"} group appears in only 16.3\% of the studies, possibly due to the difficulty in capturing emotions in-the-wild (ITW), such as the dataset discussed in Kahou et. al 
\cite{Kahou2015}, that is, when the individual is not aware of the purpose or the existence of ongoing video recording or photography. It is important to note that facial expressions do not completely correlate to what the individual is feeling, as is the case with posed facial expressions, but is generally used as an acceptable indicator for emotion, even when used in combination with other indicators \cite{WOS:000475910800001}, \cite{WOS:000807364500004}.

\subsection{ Hybrid Architectures }
Models that combine multiple networks into one architecture, called hybrid models, are becoming increasingly accurate, particularly those that combine convolutional neural networks (CNNs) and recurrent neural networks (RNNs) \cite{WOS:000518657800086} for facial emotion recognition (FER) tasks. 

In addition, recent research has shown that the integration of recurrent layers, such as the long-short term memory layer (LSTM layer) \cite{hochreiter1997long}, which processes inputs recursively, making them particularly useful for capturing the temporal dynamics of facial expressions and inserting these layers into hybrid models can further improve their performance \cite{WOS:000756540200001}. 
    
Another promising research line for improving FER accuracy is the use of multiple features such as audio and processed images in addition to facial color (RGB) images \cite{WOS:000460064700015}. These additional features can provide complementary information that can improve the robustness and accuracy of the FER system. 

However, there are still challenges that need to be addressed, such as how to effectively fuse multiple features and how to effectively train such time-consuming models. Anyhow, hybrid models with transformers or LSTMs classifiers, as well as multiple features are a promising direction for improving the state of the art in FER \cite{WOS:000756540200001}, \cite{WOS:000783834000109}.  
    
\section{Methods}
\subsection{Datasets used for Training and Prediction}
\label{subsec:Datasets}
 Considering the need for an architecture that can provide adequate accuracy and real-time response for predicting emotions in children in uncontrolled environments, we create the architecture \textit{HybridCNNFusion} to process the real-time sequence of frames. 
 
 To accomplish the task in hand, we planned on training our model on the two datasets publicly available with the highest accuracy for FER task in children \cite{WOS:000556121100052}. 
 The datasets used are the FER-2013 \cite{Sambare2022} and the Karolinska Directed Emotional Faces \cite{kdef}. Most datasets for FER tasks are aimed towards adults and with posed expressions, therefore we decided to use the ChildEFES \cite{childefes}, a private video dataset of Brazilian children posing emotions for fine-tunning.

\subsection{\textit{HybridCNNFusion} Architecture Model }
\label{subsec:Architecture}

\begin{figure}[!ht]
    \centering
    \includegraphics[width=\linewidth]{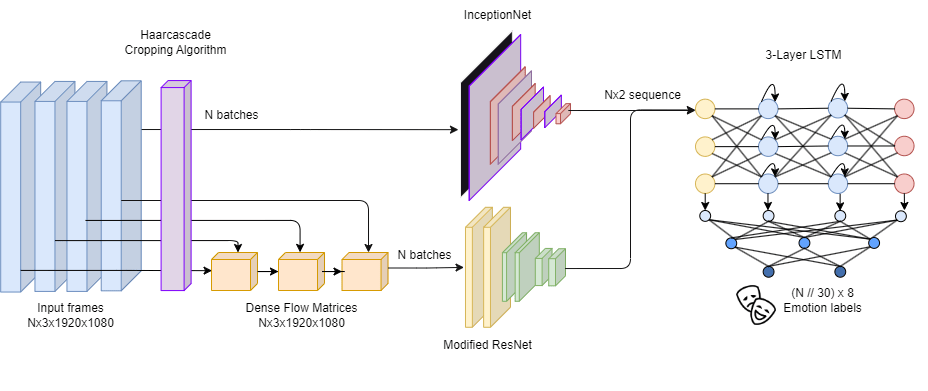}
    \caption{ \textit{HybridCNNFusion} architecture. The full implementation is available  \href{https://anonymous.4open.science/r/hybridcnnfusion-fer}{here.}}
    \label{fig:architecture}
\end{figure}

In Fig.~\ref{fig:architecture} we present the elements that make up the \textit{HybridCNNFusion} architecture. The first step in building our architecture was to allow the model to be used in in-the-wild scenarios, by implementing a Haarcascade \cite{haarcascade} region detection algorithm to center and crop the children faces. 

These cropped images are then passed to a Convolutional Neural Network (CNN), specifically the InceptionNet \cite{szegedy2014going}, to process the cropped RGB pixels generated by the Haarcascade algorithm. In parallel, we use the Gunner Farneback's algorithm \cite{optical-flow} to retrieve the dense optical flow values from the current and previous cropped frames. This is made to allow the network to process the variation in facial muscles and skin movement over time. The optical flow matrices are then passed to a second CNN, specifically a variation of the ResNet \cite{he2015deep}.

After calculating these two separate features, they are concatenated and used as input for a final recurrent block, specifically made with layers of LSTM cells to generate the concatenated intermediary output. This takes advantage of the sequential nature of the video frames to output a final vector of predicted probabilities for each emotion. The aforementioned model uses a technique called Late Fusion \cite{WOS:000756540200001}, in which two separate features are concatenated inside the architecture and used as input for the final output layers. 

The Late fusion technique allows for a better usage of the motion generated by separate facial Action Units \cite{ekman1978facial} by having to distinctly trained CNNs, one for raw RGB values (outputed by the InceptionNet) and another for dense HSV motion matrices (outputed by the ResNet + OpticalFlow combination). The use of the optical flow features as input for the ResNet allows for processing sequential information, specifically, that of motion, through clever manipulation of the raw RGB values.

The step by step used for a single video classification iteration is presented in the Algorithm~\ref{alg:ArchAlgo} section below.

\begin{algorithm}[ht]
\caption{\textit{HybridCNNFusion} pseudo algorithm}
\label{alg:ArchAlgo}
\begin{algorithmic}
\Statex \textbf{Input:} $N \times 1920 \times 1080$ RGB frames ($\Vec{x}_{i}$) and a one-hot-vector for the emotion label throughout the video ($e$).
\Statex \textbf{Output:} $E_{j}$ for each 10 second window of the frames.
\Statex \textbf{Step-by-step:}
\For{ $\Vec{e_i}, i=0:M$}
\State Cut each vector $\Vec{e_{i}}$ using the Haarcascade cropping algorithm to center the faces, 
\State to $n \times n$ sized images.
\State Apply the Gunner Farneback's algorithm to the cropped $\Vec{c_{i}}$ frames.
\State Group the cropped and optical flow features into groups of 30 frames.
\State Batch input them into two separate CNNs, respectively: 
\State CNNFlow = InceptionNet(3, 8) and CNNRaw = ResNet34(3, 8).
\EndFor
\For{$group_{j}, 0:N/30$}
\State Concatenate the cropped and optical flow features.
\State Input the concatenated vectors into a 3-layer LSTM and generate a sequence 
\State of predictions based on the previous emotion probabilities.
\State Append group emotion label to a sequence of labels for the entire video.
\EndFor
\end{algorithmic}
\end{algorithm}

\subsection{Ethical aspects and considerations of the solution}
\label{sec:Social}
The task of facial emotion recognition (FER) in children is particularly challenging due to the ethical issues and the need for a high level of precision and interpretability. 

Most existing FER approaches focus on non-ethically critical situations, such as customer satisfaction or in controlled lab conditions \cite{WOS:000483067100012}. On the other hand, the task of FER in emotionally vulnerable children requires a much greater level of trustworthiness in accordance to the ethical constraints of the psychologist-patient relationship \cite{WOS:000647216000001}. 

This specific research branch of FER tasks demands the ability to accurately detect and interpret facial expressions in real-time videos of children in in-the-wild (ITW) situations, all the while ensuring the confidence  of the information being generated \cite{WOS:000523305700001}, \cite{WOS:000447336700033}.

\section{Results}  
\label{sec:Results}
The final model implementation had memory limitations that compromised the deployment of the \textit{HybridCNNFusion} architecture. Despite this limitation, the final model was trained on both the FER2013 and KDEF datasets and fine tuned on ChildEFES dataset to maximize accuracy. The entire model could not be entirely fitted through our private dataset, so we measured partial accuracy for the intermediary models. The InceptionNet had an accuracy of about 70\%, while the ResNet had an accuracy of about 72\%. Overall, the model had a speed averaging 2.5s for a single iteration, for videos averaging 10s of duration.

The input images are cropped to the required size of both networks. The output consists of a stochastic vector of probabilities predicting one of 7 possible base emotions, as well as a neutral emotion, totaling 8 possible labels \cite{ekman1984}. Both intermediary CNNs have an output vector of size 32, and the concatenated feature is a vector with 64 entries. The final output layer has a size of $(N / 30)\times 8$, with $N / 30$ equal to the total duration of the video divided in groups of 30 frames, each group with a separate predicted emotion label.

\section{Conclusion} 
\label{sec:Conclusion}
Considering the technological aspects and the initial results obtained, the architecture proposed is a continuous push towards identifying children emotions in in-the-wild conditions, altough not yet fit for real-world usage.

The fusion of dense optical flow features in conjunction with a hybrid CNN and a recurrent model represents a promising approach in the challenging task of facial emotion recognition (FER) in children, specifically in uncontrolled environments. Being a critical need in the field of psychology, this approach offers a potential solution.

For ethically sensible situations, there are still important metrics that have to be calculated, such as the Area under the ROC Curve (AOC), which can indicate whether the model is prone to miss important emotion predictions within small and specific frames, also called micro-expressions \cite{WOS:000549823900015}.
    
In fact, there is a large gap on current ethical questions for the task, but we believe that improving the interpretability of the architecture, explainability and security of transmission of the processed information should be the focus of future models and frameworks instead of just the overall accuracy. This will ensure that the technology can be used safely and effectively to support the emotional well-being of children.

\printbibliography

\end{document}